\documentclass{article}

 \usepackage[nonatbib,final]{nips_2016}

\usepackage[utf8]{inputenc} 
\usepackage[T1]{fontenc}    
\usepackage{url}            
\usepackage{booktabs}       
\usepackage{amsfonts}       
\usepackage{nicefrac}       
\usepackage{microtype}      

\usepackage{setspace}
\usepackage[pagebackref=true,breaklinks=true,letterpaper=true,colorlinks,bookmarks=false]{hyperref}
\usepackage{epsfig}
\usepackage{graphicx}
\usepackage{amsmath}
\usepackage{amssymb}
\usepackage{subfig}
\usepackage[ruled]{algorithm}
\usepackage{algpseudocode}

\newcommand{\ud}{\,\mathrm{d}}

\newcommand{\R}{\mathbb{R}}

\newcommand{\comment}[1]{ }

\DeclareMathOperator*{\argmin}{arg\,min}
\DeclareMathOperator*{\argmax}{arg\,max}

\graphicspath{{./figures/}}

\title{Quickest Moving Object Detection}

\author{
  Dong Lao and Ganesh Sundaramoorthi\\
  King Abdullah University of Science \& Technology (KAUST), Saudi Arabia\\
  \texttt{ \{dong.lao,ganesh.sundaramoorthi\}@kaust.edu.sa}
}

\begin{document}

\maketitle

\begin{abstract}
  We present a general framework and method for simultaneous detection and segmentation of an object in a video that moves (or comes into view of the camera) at some unknown time in the video. The method is an online approach based on motion segmentation, and it operates under dynamic backgrounds caused by a moving camera or moving nuisances. The goal of the method is to detect and segment the object as soon as it moves.  Due to stochastic variability in the video and unreliability of the motion signal, several frames are needed to reliably detect the object. The method is designed to detect and segment with \emph{minimum} delay subject to a constraint on the false alarm rate. The method is derived as a problem of \emph{Quickest Change Detection}. Experiments on a dataset show the effectiveness of our method in minimizing detection delay subject to false alarm constraints.  
\end{abstract}

\section{Introduction}

Segmentation of object(s) from video is a fundamental problem in computer vision. Motion cues play a role in biological visual systems, and they may play an important part in segmentation of objects in both biological and computer vision systems \cite{ochs2014segmentation}. Thus, there have been many works that segment a video by apparent motion in an attempt to segment relevant objects (e.g., \cite{wangA94,weiss1996unified,sun2013fully,ochs2014segmentation,yang2015self}). In many of these methods, it is assumed that the video is observed when the objects of interest are already in motion and the task is to segment such objects.  However, that may not be representative of the problem solved by biological systems or that is required in certain computer vision applications such as surveillance. In such cases, the object may be stationary or out of the field of view of the observer or camera when the video starts. Thus, \emph{detection} and segmentation of the object at the time it moves is needed.

\comment{
and determine its segmentation before being tracked and segmented in subsequent frames.
}

This paper addresses the problem of \emph{detection} and \emph{segmentation} of an object in a video that moves, at some unknown time, differently than a dynamic ``background'', induced from camera motion or slight moving nuisances in the scene, such as leaves in a tree blown by wind. Since we wish to address a problem that is closer to the problem that may be solved by biological systems, an \emph{online} algorithm is desired. We define an online system as a system that receives frames sequentially, one at a time, and must make a decision, that is, declare a detection or wait for more data, at each time instant. Although observing mores frames before declaring a detection may lead to a more accurate detection and segmentation, since more motion may be observed leading to a stronger motion cue and more data mitigates the stochastic effects of background nuisances, this leads to greater delay, which may not be tolerable in certain scenarios, e.g., an antelope detecting an approaching lion, or a surveillance system detecting and tracking an intruder. Thus, our goal is to derive an algorithm with \emph{minimum delay}, defined as the number of frames acquired after the object moves. Of course one can easily achieve zero delay by always declaring detection at frame 1, irrespective of the data. Thus, we require an algorithm that operates under a constraint on \emph{false alarms}, defined as declarations of detection before the object moves or incorrect or inaccurate segmentation after the object moves.

Our specific contributions are as follows: {\bf 1.} We derive statistical models of image sequences for the preceding problem so as to formulate the problem in the framework of \emph{Quickest Change Detection (QCD)} considered in statistics \cite{poor2009quickest,veeravalli2013quickest}. That problem assumes a stochastic process, which arises from a given distribution before an \emph{unknown} change time and a different distribution post-change. That theory derives online algorithms to detect the change time with \emph{optimal} delay subject to false alarm constraints. {\bf 2.} We illustrate and address additional complications in QCD arising from moving object detection, namely, that pre- and post-change distributions depend on hidden variables (motions, occlusions, segmentations), which must be estimated from non-convex optimization problems. {\bf 3.} We derive an online algorithm for detection and segmentation of a moving object based on QCD, inheriting \emph{optimality} properties. Further, we introduce an application specific modification of the algorithm to address computational cost of the algorithm derived from QCD. {\bf 4}. Finally, we quantitatively evaluate the algorithm on a dataset and compare it to a baseline approach in terms of the desired optimality.


\subsection{Related Work}

We briefly highlight related work in motion segmentation, and change detection. Motion segmentation relies on computing apparent motion, determined from parametric models (e.g., translations or affine) of flow \cite{lucas1981iterative} or dense optical flow \cite{brox2004high,sun2010secrets,ayvaci2010occlusion}. Early works (e.g., \cite{wangA94,weiss1996unified}) on motion segmentation use parametric models of flow and solve a joint problem in segmentation and flow estimation. To deal with deforming objects such as articulating humans, non-parametric motion models of flow are assumed (e.g., \cite{sun2010layered}), and solved as a joint problem of segmentation and flow estimation. \cite{sun2013fully,yang2015self} use a similar approach to causally segment videos frame-by-frame. Those approaches typically operate on 3 frames. To obtain a stronger motion signal, whole videos are processed in batch, rather than online, in \cite{ochs2014segmentation,brox2015} by grouping trajectories of points across frames. Although the causal approach \cite{yang2015self} is online and uses few frames, the motion signal may not be strong enough to segment the object. On the other hand, \cite{ochs2014segmentation,brox2015} may achieve a stronger motion signal at the expense of processing the whole batch. Thus, an algorithm that chooses the number of frames to achieve a strong enough motion signal would be beneficial, and our approach encompasses this. Moreover, existing approaches, while addressing segmentation, do not address \emph{detection} of a moving object at an unknown time in the video, which is the main motivation of our method.

The problem of detecting changes in a video has a large literature in computer vision, too extensive to review here, and thus we refer to \cite{goyette2014novel} for a good survey. That literature mainly addresses detection and segmentation of moving objects by background subtraction \cite{benezeth2010comparative} - subtraction of a known or adaptively determined background from the current video frame. We are interested in video with moving cameras or dynamic background nuisances, for which the methods in that literature largely do not apply, although there have been advancements.  While there are methods that deal with dynamic cameras and detect and segment moving objects by motion (e.g., \cite{cohen1999detecting}), they do not address the issue of the tradeoff between detection delay and false alarms, which our approach addresses optimally by using QCD. We are not aware of the use of QCD in the detection and segmentation by motion of video. In the next section, we summarize the main ideas from QCD before framing our problem in that framework.

\comment{

proceeding to deriving our models and framing the problem of moving object detection as a QCD problem, and then deriving algorithms. Finally, we show experiments demonstrating the effectiveness of our algorithm is addressing the tradeoff between detection delay and false alarms.

\begin{itemize}

\item change detection: in nips \cite{angela2006optimal}
\item quickest change detection: \cite{poor2009quickest,veeravalli2013quickest}
\item change detection: outdated ... all are background subtraction
  \begin{itemize}
  \item frame differencing: detecting only boundaries, can accumulate over frames
  \item background subtraction: \cite{benezeth2010comparative} various background models, parametric, non-parameteric (updated over time since background model changes); sensitive to background motion; for small nuisances (leaves moving by wind), periodic background models are there; whole image can be used to compute background model via PCA (background most descriptive e-vecs for illuminative insensitivity)
  \end{itemize}
\end{itemize}
}

\section{Overview of Quickest Change Detection Theory}

\label{sec:qcd}

We introduce the main theory, \emph{Quickest Change Detection} (QCD) \cite{poor2009quickest,veeravalli2013quickest}, underpinning our algorithm for moving object detection. QCD considers the problem of \emph{sequentially} detecting changes in distribution of a discrete-time stochastic process $\{X_i \}_{i=1}^{\infty}$. It is assumed that $X_i$ is sampled from a known distribution $p_0$ before an unknown \emph{change time} $\Gamma$, and that $X_i$ is sampled from a known distribution $p_1$ at and after the change time. The literature on QCD derives algorithms for determining the change with fewest observations $X_i$ after the change subject to constraints on false alarms (a false alarm is a declared change by the algorithm before the change time $\Gamma$). The main motivation of this theory is that due to the stochastic nature of the problem, algorithms attempting detection of changes sequentially by seeking changes in adjacent samples may lead to false alarms and significant performance gains can be obtained by incorporating all samples before the current time. Of course a strategy that uses more samples would reduce false alarms, but lead to greater delay in detecting the change. QCD addresses this tradeoff between false alarms and number of samples after the change time \emph{optimally}.

QCD is formulated as an optimization problem. Based on the priors available on the change time $\Gamma$, the optimization problem can be formulated as a Bayesian or minimax problem. Since we do not have such a prior in our application, we use a minimax formulation that we review from \cite{veeravalli2013quickest} next. A \emph{stopping time} $\tau$ with respect to a stochastic process $\{X_i \}_{i=1}^{\infty}$ is a random variable such that the event $\{ \tau = n \}$ is in the sigma-algebra generated by $X_1,\ldots, X_n$. Intuitively, $\tau$ is a function that may return $n$ if it uses only information determined from $X_1,\ldots, X_n$. An example is $\tau = \inf \{ n \,:\, \sum_{i=1}^n X_i \geq b \}$, i.e., $\tau$ is the first time $n$ that the sum of $X_i$ up to time $n$ exceeds a threshold $b$. Let $\mathbb P_n$ and $\mathbb E_n$ denote the probability measure and expectation, associated with a change time of $n$. The \emph{average detection delay} of a stopping time $\tau$ in a minimax formulation is defined as 
\begin{equation}
  \mathsf{ADD}(\tau) = \sup_{n\geq 1} \mathbb E_n [ \tau - n | \tau \geq n  ]. 
\end{equation}
The delay is $\tau-n$ given that the change time is $n$ and $\tau\geq n$. $\mathsf{ADD}$ defines the worst case average delay over all change times. The \emph{false alarm rate} of a stopping time is defined as $\mathsf{FAR}(\tau) = 1 / \mathbb{E}_{\infty} [ \tau ]$, that is, one over the average stopping time given that there is no change. Note that there are different ways of defining the average detection delay and false alarm rate, but all lead to similar optimal stopping times. The QCD optimization problem is 
\begin{equation} \label{eq:stoppingrule_opt}
  \min_{\tau} \mathsf{ADD}(\tau) \mbox{ subject to } \mathsf{FAR}(\tau) \leq \alpha,
\end{equation}
where $\alpha \in [0,1]$ is the maximum tolerable false alarm rate. It can be shown that the optimal stopping time is given by a maximum likelihood ratio test:
\begin{equation}
  \tau_c = \inf\{ n \, : \, \Lambda_n \geq b \},
\end{equation}
where the threshold depends on $\alpha$, and the \emph{likelihood} $\Lambda_n$ arises from a test of the null hypothesis $\Gamma \geq n $ against the alternative hypothesis $\Gamma < n$, i.e., 
\begin{equation} \label{eq:max_likelihood}
  \Lambda_n = \mathbb{ P } [ \Gamma < n | X_1, \ldots, X_n ] / 
    \mathbb{ P } [ \Gamma \geq n | X_1, \ldots, X_n ]
   = \max_{1\leq k < n } \prod_{i=k}^n  p_1(X_i) /  p_0(X_i) , 
\end{equation}
where the last equality is made under the assumption that $X_i$ are iid before and after the change.  In our moving object detection problem that we introduce next, the post-change distribution is only known conditional on a parameter $\theta$, and in this case, the stopping time is $\Lambda_n = \max_{1\leq k < n } \max_{\theta} \prod_{i=k}^n p_{\theta}(X_i) / p_0(X_i) $, which also optimizes \eqref{eq:stoppingrule_opt}.

The sequential algorithm to detect the change is to acquire data $X_i=x_i$ and, at each new acquisition at time $n$, one computes $\Lambda_n$, by solving a maximization problem over possible change times from $k=1, \ldots, n-1$. At the first $n$ when $\Lambda_n \geq b$, a detection is declared and the maximizer $k^{\ast}_n$ is the change time. Solving the maximization problem directly may be expensive. In the case that the post-change distribution is known, previous data does not need to be revisited and $\Lambda_n$ can be computed recursively. However, the case of the post-change distribution that has an unknown parameter does not have a recursive implementation. We suggest later an approach to reduce the cost of evaluating the likelihood for all $1\leq k < n$ at each arrival of a new sample at time $n$.

\section{Quickest Moving Object Detection}

In this section, we formulate the problem of sequentially detecting and segmenting moving objects from a video quickly as a Quickest Change Detection problem. We first present our statistical models for the image sequence and moving object, then formulate them in the framework of Quickest Change, show the additional complications that arise from our models, and then propose an algorithm.

\subsection{Model for the Image Sequence and the Moving Object}

\label{subsec:models}

We assume that the scene is observed by a possibly moving observer, i.e., the ``background'' may be moving and at some time $\Gamma$ an object within the scene begins to move or comes into view of the camera. Although our algorithm will apply to more than one object that may occur at different change times, for simplicity of notation in deriviing our models, we assume the case of one moving object at one change time.

To apply Quickest Change Detection, we specify statistical models pre- and post-change. Let $\Omega \subset \R^2$ be the domain of the images, and let $I_i : \Omega\to \R^k$, $i \geq 1$ be the image sequence. We denote by $R_i \subset \Omega$ the region of the moving object at time $i \geq \Gamma$. We set $R_i = \emptyset$ for $i < \Gamma$. We denote $O_i^1 \subset R_i$ as the occlusion of the object induced by a change of viewpoint of the camera or a self-occlusion.  Similarly, $O_i^0 \subset \Omega\backslash R_i$ is the occlusion of the background induced by the previous conditions or from occlusion induced by the moving object.

The \emph{background displacement} between two adjacent frames in the image sequence is $v_i^0 : \Omega\backslash R_i\backslash O_i^0 \to \mathbb R^2$, and the \emph{object displacement} is $v_i^1 : R_i \backslash O_i^1 \to \R^2$. Although such displacements are not defined in the occluded parts of the domain, they will be smoothly extended into the occlusion. We denote the \emph{warp} between frames $i$ and $i+1$ as $w^0_{i,i+1} : \Omega\backslash R_i \backslash O_i^0  \to \Omega\backslash R_{i+1}$ and $w^1_{i,i+1} : R_i\backslash O_i^1 \to R_{i+1}$, defined by $w^j_{i,i+1}(x) = x + v^j_i(x)$, which are diffeomorphisms onto their ranges, which arise from camera motion of a non-planar scene or from deforming objects.
The warp between time 1 and time $i$ will be denoted by $w_i^j$ and can be determined from the recursion (we assume $v^1_i = 0, \, i<\Gamma$)
\begin{equation} \label{eq:sum_warps}
  w_{i+1}^j(x) = w^{j}_i(x) + v_i^j(w^{j}_i(x)), \,\, i > 1 \quad \mbox{with}\quad 
  w_{1}^j(x) = x.  
\end{equation}
The model for the evolution of the region across time after the change time is
\begin{equation} \label{eq:model_region}
  R_{i+1} = w_{i,i+1}^1(R_i) \cup D_{i+1} \mbox{ for }  i \geq \Gamma 
  \quad \mbox{and} \quad 
  R_{\Gamma} = R,
\end{equation}
where $D_{i+1} \subset \Omega$ is the part of the object that is disoccluded (comes into view of the camera) at time $i+1$, and $R$ is the region of the object at the change time $\Gamma$. Therefore, the region of the object is a smooth warping of an initial region $R$ (up to disocclusions), and therefore also smoothly varies in time.

Assuming approximate Lambertian reflectance of the scene, we may relate successive images before the change as 
\begin{equation}
  I_{i+1}(w_{i,i+1}^0(x)) = I_i(x) + \eta_i(x), \quad x\in \Omega\backslash O_i^0, \,\, i < \Gamma,
\end{equation}
and after the change time as
\begin{equation}
  \begin{cases}
    I_{i+1}(w_{i,i+1}^0(x)) = I_i(x) + \eta_i(x) & x\in \Omega \backslash
    R_i \backslash O^0_i \\
    I_{i+1}(w_{i,i+1}^1(x)) = I_i(x) + \eta_i(x) & x\in R_i\backslash O^1_i
  \end{cases}, \, i \geq \Gamma,
\end{equation}
where $\eta_i(x)$ is a Gaussian iid noise process in both $i$ and $x$, which is used to model deviations from the Lambertian assumption. We assume that $\eta_i(x) \sim \mathcal N( 0, \sigma_{\eta, 1} )$ for $x\in R_i\backslash O_i^1$ and $\eta_i(x) \sim \mathcal N( 0, \sigma_{\eta, 0} )$ for $x\in \Omega\backslash R_i\backslash O_i^0$.

\subsection{Algorithm for Detection and Segmentation}

We now apply the theory in Section~\ref{sec:qcd} to the models of the previous sub-section to derive an algorithm that detects and segments the object $R_i$ at the detection time. We must compute the likelihood ratio, $\Lambda_n$, which requires the computation of the pre- and post-change distributions. Our model is complicated by hidden variables, $R_i, O_i^j, v_i^j$, which must be marginalized out to determine the probabilities. We may treat these distributions that depend on hidden variables as a post-change distribution with an unknown parameter $\theta$, described in Section~\ref{sec:qcd}: we maximize over the hidden variables.

Let $\mathbf I_{1:n}$ denote all the images $I_1, \ldots, I_n$, and similarly define $\mathbf v_{1:n}^j$ and $\mathbf R_{1:n}$. Note that conditional on $v^j_i$ and $R_i$, the pairs of images $I_i, I_{i+1}$ are independent for all $i$. Using this, one can show that pre-change,
\begin{equation} \label{eq:p0}
  p_0(\mathbf {I}_{i:i+1} | v_i^0 ) \propto 
  \exp\left\{  -\frac{1}{2\sigma_{\eta,0}^2} \int_{\Omega} \rho( I_{i+1}(w_{i,i+1}^0(x)) - I_i(x) ) \ud x\right\},
\end{equation}
where $\rho(x) = \min\{ x^2, \beta \}$ ($\beta>0$) is a truncated quadratic, which is a robust norm that eliminates the explicit estimation of the occlusion \cite{sun2010secrets}. Similarly, the post-change conditional distribution is 
\begin{equation} \label{eq:p1}
  p_1(\mathbf {I}_{i:i+1} | v_i^0, v_i^1, R_i ) \propto
  \exp\left\{
    -\sum_{j=0}^1 \frac{1}{2\sigma_{\eta,j}^2} 
    \int_{R_i^j} \rho( I_{i+1}(w_{i,i+1}^j(x)) - I_i(x) ) \ud x
  \right\}, 
\end{equation}
where $R_i^1 = R_i$ and $R_i^0 = \Omega \backslash R_i$. By maximizing over the conditioned variables, one can use the right-most equality in \eqref{eq:max_likelihood} to deduce
\begin{equation} \label{eq:likelihood_image}
  -\log \Lambda_n = 
    \min_{1\leq k < n} \min_{ \mathbf v_{k:n}^0 } \sum_{i=k}^{n} \int_{\Omega} \mbox{Res}_i^0(x) \ud x - \min_{\mathbf{R}_{k:n} }   
  \sum_{j=0}^1 \min_{ \mathbf v_{k:n}^j } \sum_{i=k}^{n} \int_{R_i^j}
  \mbox{Res}_i^j(x) \ud x  - \log{[p(R_i)]},
\end{equation}
where $\mbox{Res}_i^j(x) = \frac{1}{2\sigma_{\eta,j}^2} \rho( I_{i+1}(w_{i,i+1}^j(x)) - I_i(x) )$.

{\bf Warp estimation}: To evaluate $\Lambda_n$, we must solve an optimization problem in the change time $k$, and the hidden variables, which we describe next. Given the other variables, we discuss the optimization in $\mathbf v_{k:n}^j$. Note that no explicit prior on the warps are assumed in the marginalization of the conditional probabilities, and thus no regularization of the warps appear in the optimization for $\Lambda_n$. Instead, we leverage on the Sobolev framework \cite{yang2015self}, to impose regularity in a natural coarse-to-fine framework allowing warps to be arbitrary diffeomorphisms by constructing warps that are a time integration of smoothly varying vector fields, which at each instant belongs to a Sobolev space. This avoids the under/over smoothing problem in global regularization typically used in optical flow.  Solving for $v_i^j$ is done as $v_i^j = \argmin_{v} \int_{R^j_i} \rho( I_{i+1}(x+v(x)) -I_i(x) ) \ud x$, as other terms are independent of $v_i^j$.

{\bf Region estimation}: Given estimates of the warps, we optimization for $\mathbf{R}_{k:n}$, which involves the second summation in \eqref{eq:likelihood_image}. Note that $R_i$ are coupled through \eqref{eq:model_region}. In the case of no disocclusions, optimizing in $\mathbf R_{k:n}$ can be replaced by optimization in any one region $R_m$ with $k\leq m \leq n$ subject to the constraint that the other regions are warps of $R_m$. This gives the minimizer of $\Lambda_n$ as the minimizer of 
\begin{equation}
  E(R_m) = \sum_{j=0}^1 \int_{R^j_m}  f_i^j(x) \ud x - \log{[p(R_m)]}, \quad
  f^j(x) = \sum_{i=k}^n \rho( I_{i+1}(w^j_{m,i+1}(x)) - I_{i}(w^j_{m,i}(x)) ),
\end{equation}
where $w^j_{m,i}$ denotes the warp of $R^j_m$ to $R^j_i$, determined by \eqref{eq:sum_warps} using given estimates of $\mathbf v_{k:n}^j$. Due to the aperture problem and occlusions, motion residuals in these cases are unreliable to estimate $R_m$, and we rely on image statistics as in \cite{yang2015self}. Thus, we modify the energy as 
\begin{equation} \label{eq:energy_motion_seg}
  E_{seg}(R_m) = \sum_{j=0}^1 \int_{R^j_m} (1-\mbox{maf}(x)) f^j(x) - \mbox{maf}(x) \log p_{R^j_m}(I_m(x)) \ud x - \log{[p(R_m)]},
\end{equation}
where $\mbox{maf} : \Omega\to [0,1]$ is the \emph{motion ambiguity function}, and $p_{R^j_m}$ are local color histograms of $I_m$ within regions. The motion ambiguity function is 1 if pixel $x\in R_m$ is occluded in \emph{all} frames $i=k,\ldots, n, i\neq m$ (all high residuals exceed $\beta$), or if $x$ is in a textureless sub-region of $I_m$. In these cases, motion cues are unreliable, and thus we rely on color histograms to group $x$. We choose the prior $p(R_m)$ to induce smoothness, as is typical in segmentation problems. The energy \eqref{eq:energy_motion_seg} now fits into a form considered in \cite{yang2015self}. Thus, we use the optimization specified there, which uses gradient descent due to non-convexity. Even though we assumed no disocclusions, optimization of \eqref{eq:energy_motion_seg} implicitly computes disocclusion as part of the grouping procedure. In particular, disocclusions (parts of the object that come into view) at time $m$ are assumed to be parts of the image moving similar to $R_m$ (or similar color intensity in the case of motion unreliability). Note that the although the optimization problem is in $R_m$, each of $R_i$ for $i\in \{k,\ldots,n\} \backslash \{ m \}$ can be determined as a propagation of $R_m$ through the sequence determined by warps $w^j_{i,i+1}$, which when done through \cite{yang2015self}, includes disocclusion in the $R_i$.

{\bf Joint Region and Warp Estimation}: For a given change time $k$ and time $n$, we state our algorithm to estimate the detected region $R_{k,n}$ in frame $n$, as well as the inner optimization in \eqref{eq:likelihood_image}, which we denote $\log\Lambda_{k,n}$. Although we only seek $R_{k,n}$ and $\Lambda_{k,n}$, we must estimate the nuisance variables $\mathbf v^j_{k:n}$ and $\mathbf R_{k:n}$. The algorithm is stated in Algorithm~\ref{alg:region_segmentation}. The gradient descent of $E_{seg}$ requires an initialization for $R_m$. We initialize it with a clustering of the cumulative optical flow from frame $m$ to $n$ (and frame $m$ to $k$). Both forward and backward direction warps are used to address incorrect grouping due to occlusion.  Because accurate warp estimation requires a segmentation, which is unknown, we use Classic-NL \cite{sun2010secrets}, robust to motion discontinuities, to roughly approximate the warps defined on $\Omega$ without a segmentation.

\begin{algorithm}
  \begin{algorithmic}[1]
    \State // {\it initialize $R_m$ for gradient descent of $E_{seg}$ }
    \State Compute the Classic-NL warp $w_{i,i+1}^{NL} : \Omega\to\Omega$
    \State Compute $w_{m,n}^{NL}, w_{m,k}^{NL}$ using \eqref{eq:sum_warps}
    and cluster together into two regions, denoted $R_m^j, j=0,1$
    \Repeat{ // {\it gradient descent of $E_{seg}$ for region $R_m$} }
    \State Compute $R_i$ by propagating $R_m$ frame by frame to frame $i$ using \cite{yang2015self}
    \State Solve for Sobolev warp, $w_{i,i+1}^j = \argmin_{w} \int_{R^j_i} \mbox{Res}^j_i(x) \ud x$, extended to $\Omega$
    \State Compute $w^{j}_{m,i}$ for $i=k,\ldots, n$ using \eqref{eq:sum_warps}
    \State Compute $f^j$ and update $R_m$ by a gradient step of $E_{seg}$ in \eqref{eq:energy_motion_seg}
    \Until{ $R_m$ does not change between iterations }
    \State Compute Sobolev warps $w^{null}_{i,i+1} : \Omega\to\Omega$ for null hypothesis
    \State Set $\mbox{Res}^{null}_i(x) = \rho( I_{i+1}(w^n_{i,i+1}(x)) - I_i(x) )$ and 
    $\mbox{Res}^j_i(x) = \rho( I_{i+1}(w^j_{i,i+1}(x)) - I_i(x) )$
    \State Compute $-\log\Lambda_{k,n} = \sum_{i=k}^n \int_{\Omega} \mbox{Res}^{null}_i(x)\ud x - \sum_{j=0}^1\int_{R^j_i} \mbox{Res}^j_i(x) \ud x$
    \State \Return $\Lambda_{k,n}$ and $R_n$ as the proposed detection in frame $n$
  \end{algorithmic}
  \caption{\sl Region segmentation and likelihood $\Lambda_{k,n}$ given change time $k$ and time $n$.}
  \label{alg:region_segmentation}
\end{algorithm}

{\bf Object Detection}: Algorithm~\ref{alg:quickest_m_object} specifies Quickest Moving Object Detection. The algorithm re-estimates $\Lambda_n$ online at each new arrival of $I_n$. At each new acquisition of $I_n$, the algorithm finds a change time $k\in \{2, \ldots, n-1\}$ by revisiting each previous data and solving an optimization problem (Lines 4-6), which is solved by Algorithm~\ref{alg:region_segmentation}.

\begin{algorithm}
  \begin{algorithmic}[1]
    \State Set $n=1$
    \Repeat{ // {\it compute likelihood ratio $\Lambda_0$  } }
    \State  Increment $n \leftarrow n+1$, acquire image $I_n$
    \For { $k = 2, \ldots, n-1$ } // {\it find a most probable change time $k$} 
    \State Compute $\Lambda_{k,n}$ and $R_{k,n}$ using Algorithm~\ref{alg:region_segmentation}
    \EndFor
    \State Set $\Lambda_n = \Lambda_{k_n^{\ast},n}$ and $R_n = R_{k_n^{\ast}, n}$ where $k_n^{\ast} = \argmax_{2 \leq k <n} \Lambda_{k,n}$ 
    \Until{ $\Lambda_n > b$ }
    \State \Return $R_n$ as the object detection at time $n$
  \end{algorithmic}
  \caption{\sl Quickest Moving Object Detection and Segmentation}
  \label{alg:quickest_m_object}
\end{algorithm}

{\bf Simplification for Efficiency}: Since the post-change distribution depends on hidden variables, data must be revisited as discussed in Section~\ref{sec:qcd}. Thus, in the general case, Lines 4-6 in Algorithm~\ref{alg:quickest_m_object} cannot be simplified without possible loss of optimality. However, we propose a heuristic that estimates the change time $k_n^{\ast}$ without having to explicitly evaluate $\Lambda_{k,n}$ for each $k$. This is done by applying QCD to simpler distributions than \eqref{eq:p0} and \eqref{eq:p1}. If the residuals $\mbox{Res}^{NL}_i(x)$ are assumed iid and distributed according to $\mathcal N(\mu_0, \sigma_0)$ pre-change and $\mathcal N(\mu_1, \sigma_1)$ post-change, then the following statistic, which can be computed efficiently, arises
\begin{equation} \label{eq:F}
  F_{k,n} = 
  \sum_{i=k}^n \exp\left\{ 
    -\frac{1}{|\Omega|}\int_{\Omega}  
    \frac{(\mbox{Res}^{NL}_i(x) - \hat \mu_{1,k})^2}{ \hat \sigma_{1,k}^2}  - 
    \frac{(\mbox{Res}^{NL}_i(x) - \hat \mu_{0,k})^2}{ \hat \sigma_{0,k}^2}
    \ud x
  \right\},
\end{equation}
where $\hat \mu_{0,k}, \hat \sigma_{0,k}$ are estimated from $\mbox{Res}_i^{NL}(x), i=1,\ldots,k$ and $\hat \mu_{1,k}, \hat \sigma_{1,k}$ are estimated from $\mbox{Res}_i^{NL}(x), i=k+1,\ldots,n$. Therefore, we modify Algorithm~\ref{alg:quickest_m_object} to Algorithm~\ref{alg:f_quickest_m_object}. Note that the $k^{\ast}$ that achieves a maximum of $F_{k,n}$ is proposed as the maximizer of $\Lambda_{k,n}$. Only $\Lambda_{k^{\ast},n}$, which approximates $\Lambda_n$ and saves computation. Note that $F_{k^{\ast},n}$ is not thresholded to obtain the stopping time. Although this would lead to additional computational savings, this is avoided since it is difficult to tune a threshold for a wide range of sequences. This is expected since the distributions that $F_{k,n}$ assumes are not as accurate as \eqref{eq:p0} and \eqref{eq:p1}. However, the maximizer $k^{\ast}$ is a useful candidate for the change time in estimating $\Lambda_{n}$.

\begin{algorithm}
  \begin{algorithmic}[1]
    \State Set $n=1$
    \Repeat{ // {\it compute likelihood ratio $\Lambda_0$  } }
    \State  Increment $n \leftarrow n+1$, acquire image $I_n$
    \For { $k = 2, \ldots, n-1$ }  // {\it find a most probable change time $k$} 
    \State Compute $F_{k,n}$ as in \eqref{eq:F}
    \EndFor
    \State Set $k_n^{\ast} = \argmax_{2 \leq k <n} F_{k,n}$,
    compute $\Lambda_n = \Lambda_{k_n^{\ast},n}$ and $R_n = R_{k_n^{\ast}, n}$ using Algorithm~\ref{alg:region_segmentation}
    \Until{ $\Lambda_n > b$ }
    \State \Return $R_n$ as the object detection at time $n$
  \end{algorithmic}
  \caption{\sl Faster Quickest Moving Object Detection and Segmentation}
  \label{alg:f_quickest_m_object}
\end{algorithm}

\comment{
We introduce one further heuristic that induces a computational cost savings. If the likelihood ratio test $\Lambda_n$ fails, and $k^{\ast}_n$ was the estimated change time at time $n$, then we 

the likelihood ratio test at a future time would pass unless the estimated change time $k^{\ast}_{n'}$ ($n'>n$)  differs 
}

\section{Experiments}

We now test our method in experiments and verify it against the theory of Quickest Change. Since we are not aware of an extensive dataset for the segmentation and detection of moving objects, which start to move or come into view at some unknown time in the video and continues moving, against a possibly dynamic scene or moving camera, we collected our own. We have collected 20 videos, some that we collected from Youtube, and a number from the dataset \cite{goyette2014novel}, those that contain camera motion or dynamic backgrounds. Ground truth segmentations are available for the videos in \cite{goyette2014novel}, and we segmented the others. Each sequence contains 100-200 frames.

We are not aware of a competing online detection and segmentation approach from video that addresses detection with minimum delay, operating with false alarm constraints. According to the theory, the derived algorithm provides the optimal solution under the models assumed in Section~\ref{subsec:models}. However, in practice due to the estimation of probabilities, which depend on hidden variables that are determined by local optimization methods, the algorithm may not be optimal. However, we show next that compared to a baseline approach of quickest change detection determined from thresholding the test statistic $F_{k,n}$ \eqref{eq:F}, our method achieves smaller delay for any false alarm constraint.

\begin{figure}
  \centering
  \includegraphics[clip,trim=20 0 25 0,width=0.33\textwidth]{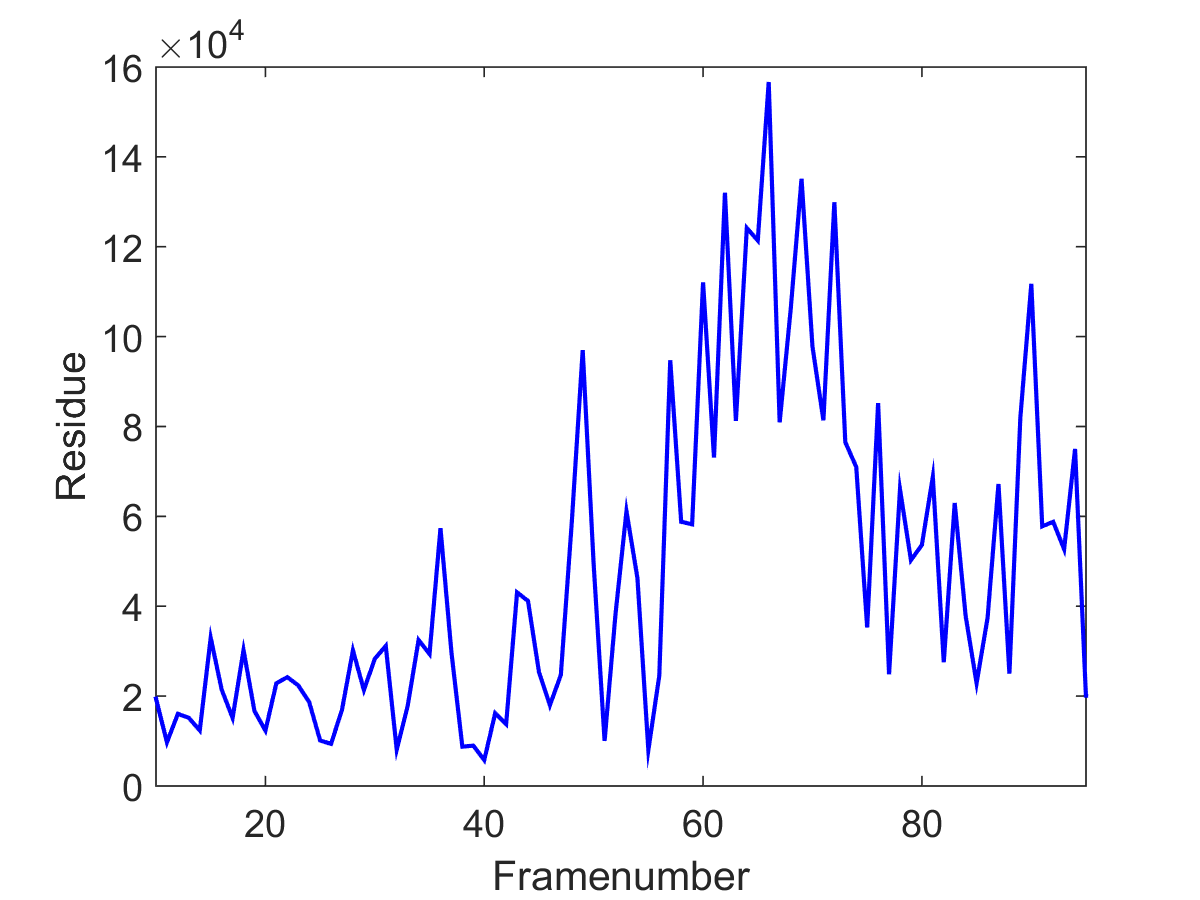}%
  \includegraphics[clip,trim=20 0 25 0,width=0.33\textwidth]{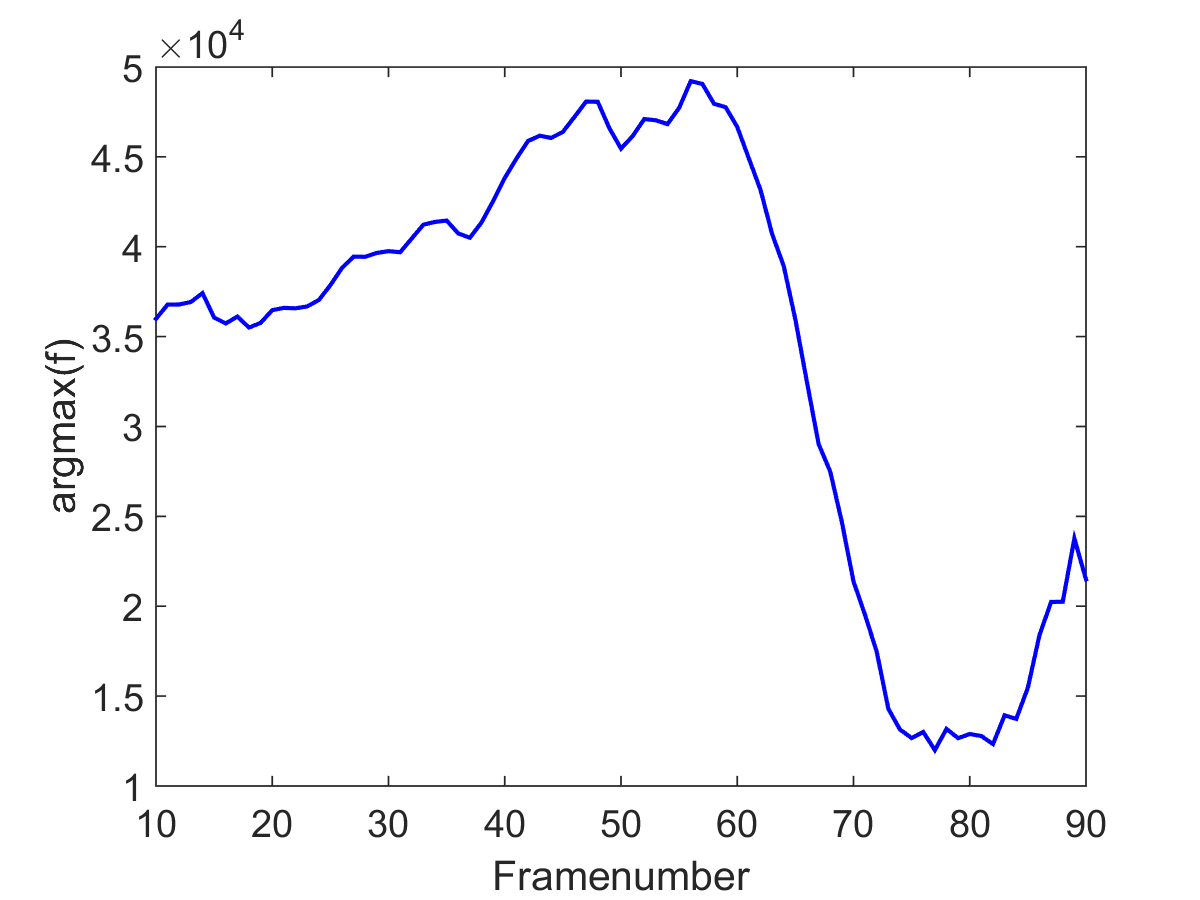}%
  \includegraphics[clip,trim=20 0 25 0,width=0.33\textwidth]{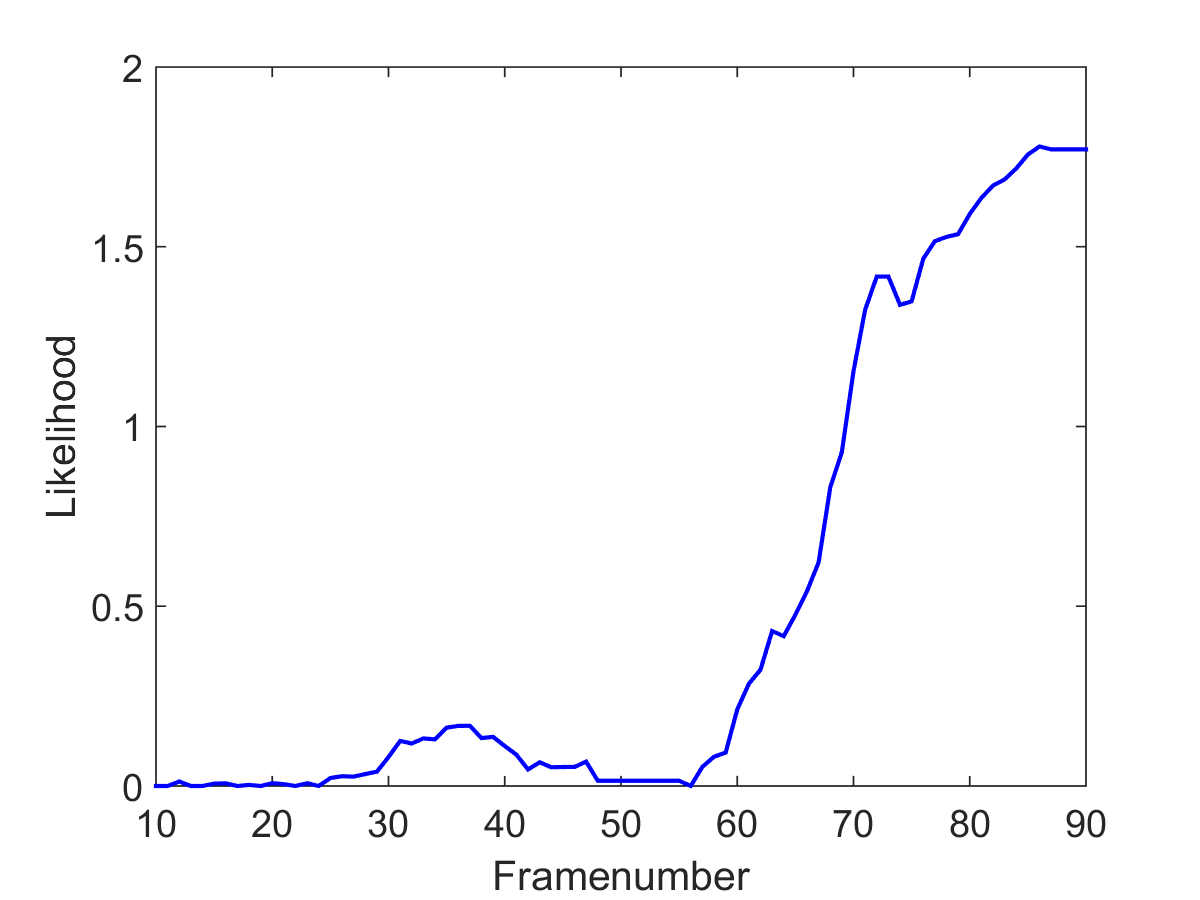}%
  \caption{ \sl\small{\bf Time evolution of statistics}. We examine a image sequence where change happens at $\Gamma=58$.  [Left]: A plot of the mean value of the residual $\mbox{Res}^{NL}_i$ from Classic-NL optical flow \cite{sun2010secrets} versus frame number in the video, which is too noisy for detecting changes. [Middle]: A plot of the statistic $F_{k^{\ast}_n, n}$ versus the frame number $n$. This statistic is less noisy, but unreliable for thresholding across sequences. [Right]: The likelihood ratio $\Lambda_n$ versus frame number $n$, which can be thresholded over a wide range to achieve correct detection.}
  \label{fig:statistic_plot}
\end{figure}

First, we illustrate the advantage of considering the likelihood ratio $\Lambda_n$ over simpler statistics in Figure~\ref{fig:statistic_plot}. There, the mean value of the frame-by-frame residual from Classic-NL is shown, which illustrates the difficulty in detecting the change due to the noise in the statistics. The plot of $\Lambda_n$ versus time $n$ shows a clean signal, and confirms our expected behavior of growing over after the change when more evidence is available. It can easily be thresholded over a wide range without false alarm. The simpler statistic $F_{k,n}$ is cleaner that the residual, but many times does not show the desired behavior, although its optimizer $k^{\ast}_n$ (not shown in the plot), is useful in computing $\Lambda_n$.

\begin{figure}
  \centering
  \includegraphics[width=0.4\textwidth]{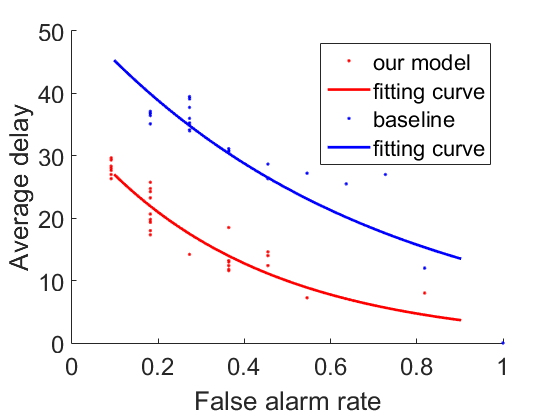}
  \caption{\sl\small {\bf  Delay versus false alarm tradeoff}.  The average detection delay (over all sequences in our dataset) versus the false alarm rate as the threshold is varied. The blue curve uses thresholds of $F_{k^{\ast}_n, n}$ as a stopping time. The red curve uses our proposed stopping time from $\Lambda_n$. Our approach leads to less delay.}
  \label{fig:ADD_versus_FAR}
\end{figure}

\begin{figure}
  \centering
  {\small
    Column: Sequence shown at frame of object detection; Row: Result for given threshold in likelihood test
  \begin{minipage}[b]{1\textwidth}
    \begin{tabular}{c@{\hskip 0.02in}c@{\hskip 0.02in}c@{\hskip 0.02in}c@{\hskip 0.02in}c@{\hskip 0.02in}c@{\hskip 0.02in}c@{\hskip 0.02in}c}
    \rotatebox{90}{\,\,$b=0.1$}
    \includegraphics[width=0.112\textwidth]{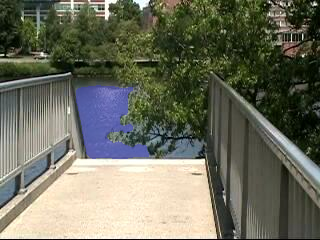} &
    \includegraphics[width=0.112\textwidth]{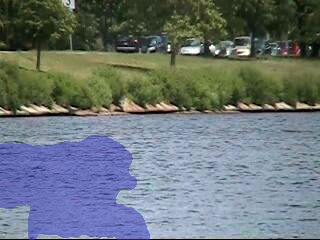}             &
    \includegraphics[width=0.112\textwidth]{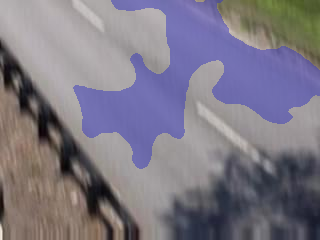} &
    \includegraphics[width=0.112\textwidth]{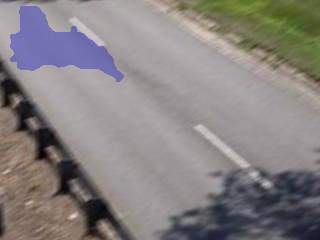} &
    \includegraphics[width=0.112\textwidth]{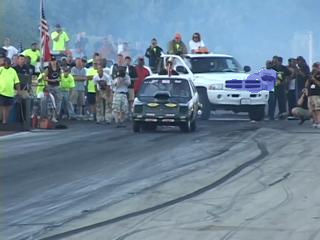} &
    \includegraphics[width=0.112\textwidth]{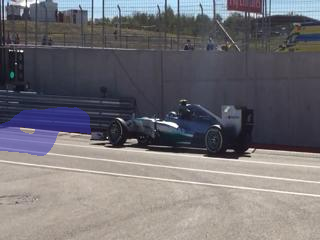} &
    \includegraphics[width=0.112\textwidth]{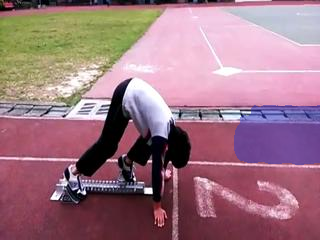} &
    \includegraphics[width=0.112\textwidth]{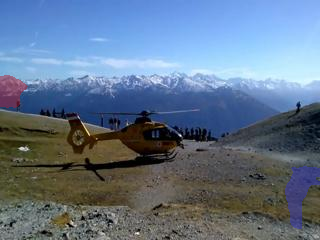} \\
      \rotatebox{90}{\,\,$b=0.4$}
    \includegraphics[width=0.112\textwidth]{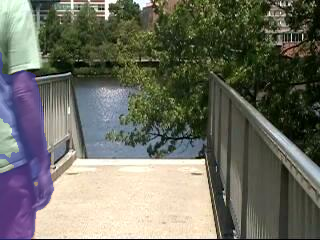} &
     \includegraphics[width=0.112\textwidth]{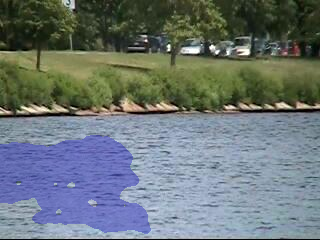}             &
    \includegraphics[width=0.112\textwidth]{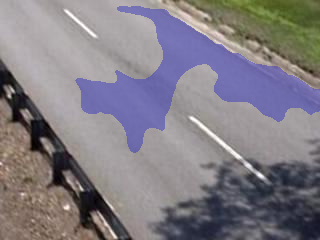} &
    \includegraphics[width=0.112\textwidth]{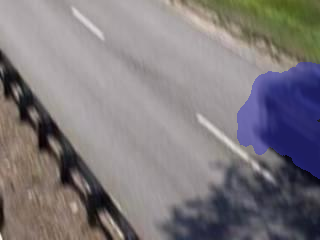} &
    \includegraphics[width=0.112\textwidth]{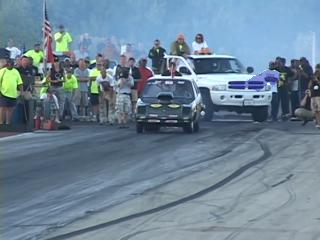} &
    \includegraphics[width=0.112\textwidth]{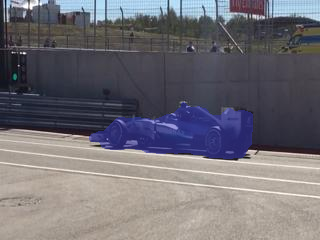} &
    \includegraphics[width=0.112\textwidth]{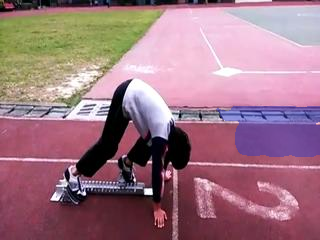} &
    \includegraphics[width=0.112\textwidth]{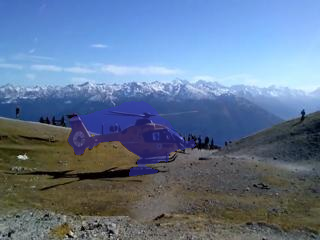} \\
    \rotatebox{90}{\,\,$b=0.7$}
    \includegraphics[width=0.112\textwidth]{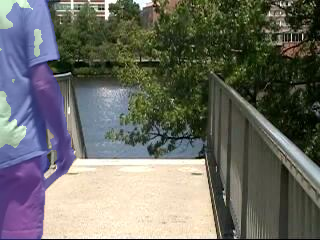} &
    \includegraphics[width=0.112\textwidth]{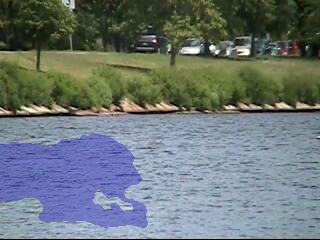}             &
    \includegraphics[width=0.112\textwidth]{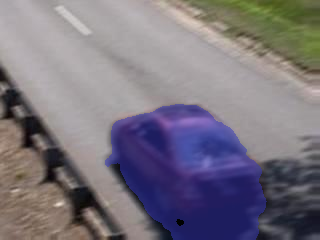} &
    \includegraphics[width=0.112\textwidth]{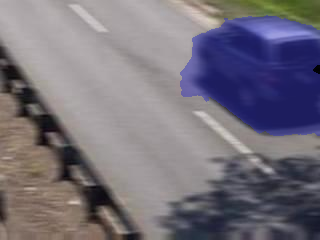} &
    \includegraphics[width=0.112\textwidth]{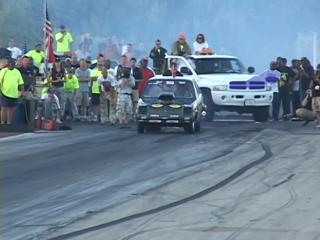} &
    \includegraphics[width=0.112\textwidth]{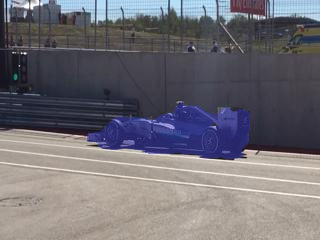} &
    \includegraphics[width=0.112\textwidth]{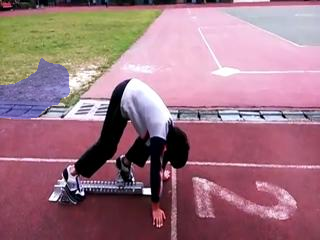} &
    \includegraphics[width=0.112\textwidth]{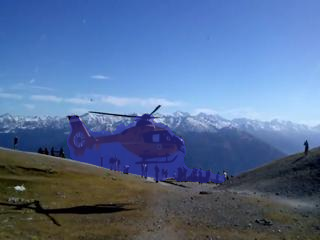} \\
    \rotatebox{90}{\,\,$b=1.0$}
    \includegraphics[width=0.112\textwidth]{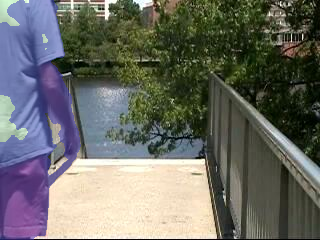} &
     \includegraphics[width=0.112\textwidth]{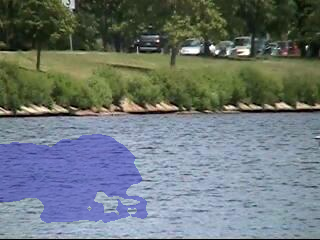}             &
    \includegraphics[width=0.112\textwidth]{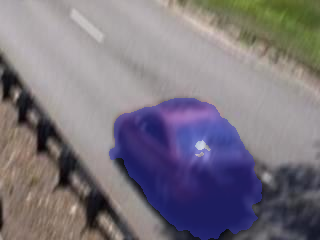} &
    \includegraphics[width=0.112\textwidth]{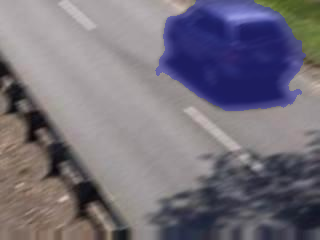} &
    \includegraphics[width=0.112\textwidth]{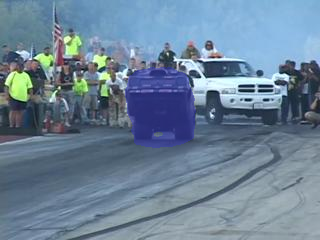} &
    \includegraphics[width=0.112\textwidth]{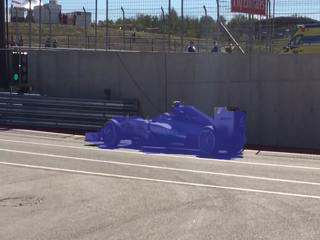} &
    \includegraphics[width=0.112\textwidth]{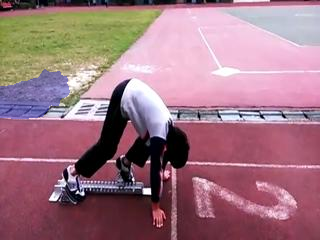} &
    \includegraphics[width=0.112\textwidth]{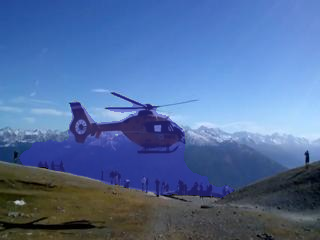} \\
      \rotatebox{90}{\,\,$b=1.3$}
    \includegraphics[width=0.112\textwidth]{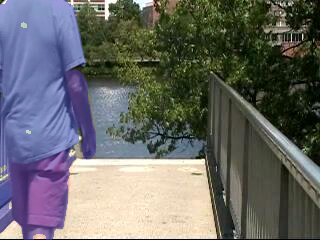} &
     \includegraphics[width=0.112\textwidth]{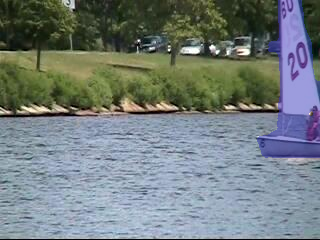}             &
    \includegraphics[width=0.112\textwidth]{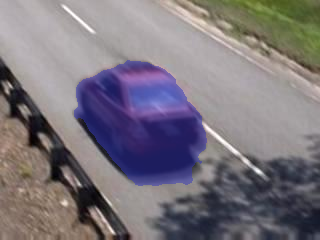} &
    \includegraphics[width=0.112\textwidth]{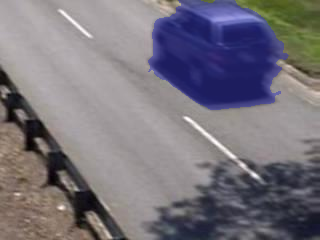} &
    \includegraphics[width=0.112\textwidth]{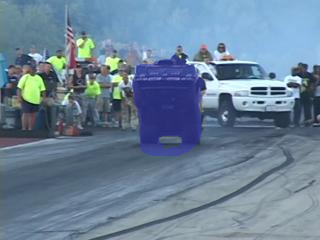} &
    \includegraphics[width=0.112\textwidth]{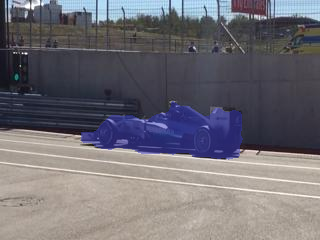} &
    \includegraphics[width=0.112\textwidth]{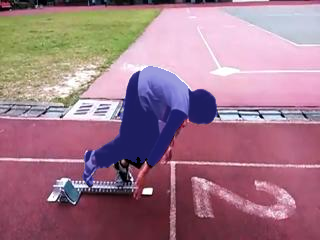} &
    \includegraphics[width=0.112\textwidth]{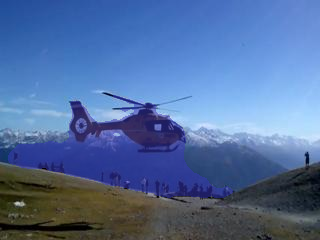} \\
    \rotatebox{90}{\,\,$b=1.6$}
    \includegraphics[width=0.112\textwidth]{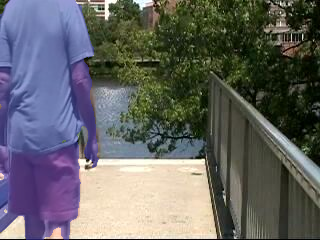} &
   \includegraphics[width=0.112\textwidth]{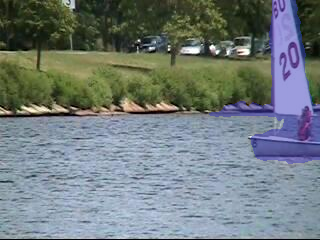}             &
    \includegraphics[width=0.112\textwidth]{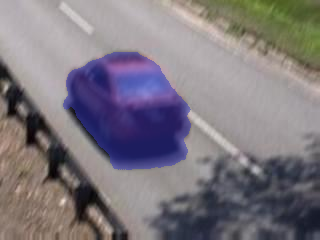} &
    \includegraphics[width=0.112\textwidth]{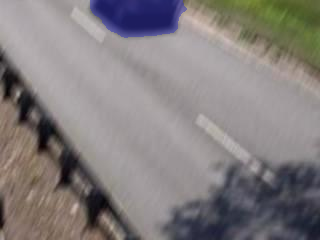} &
    \includegraphics[width=0.112\textwidth]{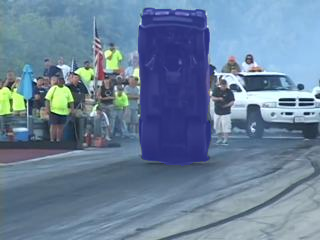} &
    \includegraphics[width=0.112\textwidth]{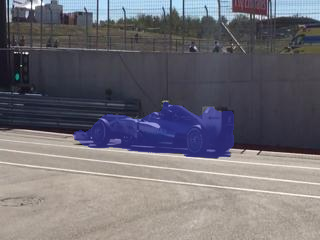} &
    \includegraphics[width=0.112\textwidth]{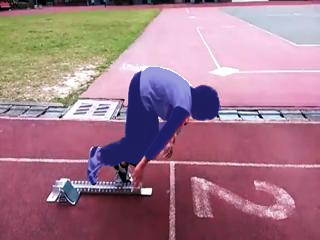} &
    \includegraphics[width=0.112\textwidth]{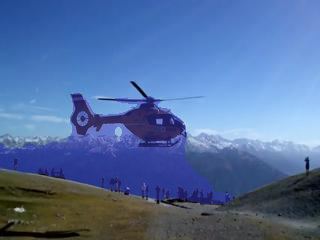} \\
    \end{tabular}
  \end{minipage}
}
\caption{\sl\small {\bf Sample detections for varying thresholds of $\Lambda_n$}. Each sequence contains one object (person, boat, car, or helicopter) moving at some unknown time with camera motion and dynamic background. Segmentations are purple masks.
  Small thresholds lead to several false alarms. As the threshold is increased, false alarms are mitigated, but with greater delay (number of frames after the object moves). The right column is a failure case, which occurs since the segmentation gets trapped in irrelevant background clutter.}
\label{fig:sample_detections}
\end{figure}

Next we run our method over the whole dataset and different thresholds $b=0.1, \ldots, 1.8$. Sample qualitative results of the segmentations and detections are shown in Figure~\ref{fig:sample_detections}. It generally shows fewer false alarms as the threshold increases, but with greater delay (when, for example, more of the object moves within the video). We quantity the results using a plot of the false alarm rate versus the average detection delay in Figure~\ref{fig:ADD_versus_FAR}. The average detection delay is defined empirically as the average difference of the detection time and the ground truth change time, i.e., $n-\Gamma$ (if $n\geq \Gamma$) over all sequences. The empirical false alarm rate is the number of detections before the change time or segmentation (with less than 0.75 f-measure) after the change time divided by the number of sequences. The figure shows the trend of fewer false alarms at the cost of longer delay. Also, we show the results of a baseline change detection procedure using a thresholding of $F_{k,n}$, which has greater detection delay than our approach at any false alarm rate.

\section{Conclusion}

We have derived a method for moving object detection and segmentation in video under the assumption of dynamic backgrounds. The method was derived based on Quickest Change Detection, which led us to an online algorithm that minimizes the delay in detection subject to false alarm constraints. Although our problem leads to a non-recursive algorithm, where the data must be revisited at arrival of new data, we proposed a simplifying heuristic that led to cost savings and was used effectively in experiments. The experiments showed that our algorithm achieved less delay under various false alarm constraints compared to a baseline approach. Future work will look into further speed-ups.

\bibliographystyle{IEEEtran}
\bibliography{detection}

\begin{thebibliography}{10}
\providecommand{\url}[1]{#1}
\csname url@samestyle\endcsname
\providecommand{\newblock}{\relax}
\providecommand{\bibinfo}[2]{#2}
\providecommand{\BIBentrySTDinterwordspacing}{\spaceskip=0pt\relax}
\providecommand{\BIBentryALTinterwordstretchfactor}{4}
\providecommand{\BIBentryALTinterwordspacing}{\spaceskip=\fontdimen2\font plus
\BIBentryALTinterwordstretchfactor\fontdimen3\font minus
  \fontdimen4\font\relax}
\providecommand{\BIBforeignlanguage}[2]{{%
\expandafter\ifx\csname l@#1\endcsname\relax
\typeout{** WARNING: IEEEtran.bst: No hyphenation pattern has been}%
\typeout{** loaded for the language `#1'. Using the pattern for}%
\typeout{** the default language instead.}%
\else
\language=\csname l@#1\endcsname
\fi
#2}}
\providecommand{\BIBdecl}{\relax}
\BIBdecl

\bibitem{ochs2014segmentation}
P.~Ochs, J.~Malik, and T.~Brox, ``Segmentation of moving objects by long term
  video analysis,'' \emph{PAMI}, vol.~36, no.~6, pp. 1187--1200, 2014.

\bibitem{wangA94}
J.~Y. Wang and E.~H. Adelson, ``Representing moving images with layers,''
  \emph{IEEE TIP}, vol.~3, no.~5, pp. 625--638, 1994.

\bibitem{weiss1996unified}
Y.~Weiss and E.~H. Adelson, ``A unified mixture framework for motion
  segmentation: Incorporating spatial coherence and estimating the number of
  models,'' in \emph{Computer Vision and Pattern Recognition, 1996. Proceedings
  CVPR'96, 1996 IEEE Computer Society Conference on}.\hskip 1em plus 0.5em
  minus 0.4em\relax IEEE, 1996, pp. 321--326.

\bibitem{sun2013fully}
D.~Sun, J.~Wulff, E.~B. Sudderth, H.~Pfister, and M.~J. Black, ``A
  fully-connected layered model of foreground and background flow,'' in
  \emph{CVPR}.\hskip 1em plus 0.5em minus 0.4em\relax IEEE, 2013, pp.
  2451--2458.

\bibitem{yang2015self}
Y.~Yang, G.~Sundaramoorthi, and S.~Soatto, ``Self-occlusions and disocclusions
  in causal video object segmentation,'' in \emph{Proceedings of the IEEE
  International Conference on Computer Vision}, 2015, pp. 4408--4416.

\bibitem{poor2009quickest}
H.~V. Poor and O.~Hadjiliadis, \emph{Quickest detection}.\hskip 1em plus 0.5em
  minus 0.4em\relax Cambridge University Press Cambridge, 2009, vol.~40.

\bibitem{veeravalli2013quickest}
V.~V. Veeravalli and T.~Banerjee, ``Quickest change detection,'' \emph{Academic
  press library in signal processing: Array and statistical signal processing},
  vol.~3, pp. 209--256, 2013.

\bibitem{lucas1981iterative}
B.~D. Lucas, T.~Kanade \emph{et~al.}, ``An iterative image registration
  technique with an application to stereo vision.'' in \emph{IJCAI}, vol.~81,
  1981, pp. 674--679.

\bibitem{brox2004high}
T.~Brox, A.~Bruhn, N.~Papenberg, and J.~Weickert, ``High accuracy optical flow
  estimation based on a theory for warping,'' in \emph{ECCV}.\hskip 1em plus
  0.5em minus 0.4em\relax Springer, 2004, pp. 25--36.

\bibitem{sun2010secrets}
D.~Sun, S.~Roth, and M.~J. Black, ``Secrets of optical flow estimation and
  their principles,'' in \emph{CVPR}.\hskip 1em plus 0.5em minus 0.4em\relax
  IEEE, 2010, pp. 2432--2439.

\bibitem{ayvaci2010occlusion}
A.~Ayvaci, M.~Raptis, and S.~Soatto, ``Occlusion detection and motion
  estimation with convex optimization,'' in \emph{Advances in neural
  information processing systems}, 2010, pp. 100--108.

\bibitem{sun2010layered}
D.~Sun, E.~B. Sudderth, and M.~J. Black, ``Layered image motion with explicit
  occlusions, temporal consistency, and depth ordering,'' in \emph{Advances in
  Neural Information Processing Systems}, 2010, pp. 2226--2234.

\bibitem{brox2015}
T.~B. M.~Keuper, B.~Andres, ``Motion trajectory segmentation via minimum cost
  multicuts,'' \emph{IEEE International Conference on Computer Vision (ICCV)},
  pp. 3271--3279, 2015.

\bibitem{goyette2014novel}
N.~Goyette, P.-M. Jodoin, F.~Porikli, J.~Konrad, and P.~Ishwar, ``A novel video
  dataset for change detection benchmarking,'' \emph{Image Processing, IEEE
  Transactions on}, vol.~23, no.~11, pp. 4663--4679, 2014.

\bibitem{benezeth2010comparative}
Y.~Benezeth, P.-M. Jodoin, B.~Emile, H.~Laurent, and C.~Rosenberger,
  ``Comparative study of background subtraction algorithms,'' \emph{Journal of
  Electronic Imaging}, vol.~19, no.~3, pp. 033\,003--033\,003, 2010.

\bibitem{cohen1999detecting}
I.~Cohen and G.~Medioni, ``Detecting and tracking moving objects for video
  surveillance,'' in \emph{Computer Vision and Pattern Recognition, 1999. IEEE
  Computer Society Conference on.}, vol.~2.\hskip 1em plus 0.5em minus
  0.4em\relax IEEE, 1999.

\end{thebibliography}

\end{document}